\documentclass{article} % For LaTeX2e
\usepackage{iclr2025_conference,times}

\usepackage{amsmath,amsfonts,bm}

\def\eqref#1{equation~\ref{#1}}
\def\1{\bm{1}}

\DeclareMathAlphabet{\mathsfit}{\encodingdefault}{\sfdefault}{m}{sl}
\SetMathAlphabet{\mathsfit}{bold}{\encodingdefault}{\sfdefault}{bx}{n}

\usepackage{hyperref}
\usepackage{url}
\usepackage{graphicx} 
\usepackage{wrapfig}

\vspace{-0.2cm}
\title{Augmenting Image Annotation: A Human– LMM Collaborative Framework for Efficient Object Selection and Label Generation}

\vspace{-0.2cm}
\author{He Zhang \\
College of Information Sciences and Technology\\
Pennsylvania State University\\
University Park, PA 16802, USA \\
\texttt{hpz5211@psu.edu} \\
\And
Xinyi Fu\thanks{Corresponding Author} \\
The Future Laboratory \\
Tsinghua University \\
Beijing, China \\
\texttt{fuxy@tsinghua.edu.cn} \\
\AND
John M. Carroll  \\
College of Information Sciences and Technology\\
Pennsylvania State University\\
University Park, PA 16802, USA \\
\texttt{jmc56@psu.edu}
}

\iclrfinalcopy % Uncomment for camera-ready version, but NOT for submission.

\begin{document}

\maketitle

\vspace{-0.8cm}

\begin{abstract}
\vspace{-0.2cm}
Traditional image annotation tasks rely heavily on human effort for object selection and label assignment, making the process time-consuming and prone to decreased efficiency as annotators experience fatigue after extensive work. This paper introduces a novel framework that leverages the visual understanding capabilities of large multimodal models (LMMs), particularly GPT, to assist annotation workflows. In our proposed approach, human annotators focus on selecting objects via bounding boxes, while the LMM autonomously generates relevant labels. This human-AI collaborative framework enhances annotation efficiency by reducing the cognitive and time burden on human annotators. By analyzing the system's performance across various types of annotation tasks, we demonstrate its ability to generalize to tasks such as object recognition, scene description, and fine-grained categorization. Our proposed framework highlights the potential of this approach to redefine annotation workflows, offering a scalable and efficient solution for large-scale data labeling in computer vision. Finally, we discuss how integrating LMMs into the annotation pipeline can advance bidirectional human–AI alignment, as well as the challenges of alleviating the ``endless annotation'' burden in the face of information overload by shifting some of the work to AI.
\end{abstract}

\vspace{-0.6cm}

\section{Introduction}
\vspace{-0.2cm}
With the rapid development of deep learning and big data technologies, image annotation, an essential component in computer vision tasks has found widespread applications in fields~\citep{najafabadi2015deep} such as autonomous driving~\citep{Huang_2018_CVPR_Workshops}, intelligent surveillance~\citep{dharmawan2022region}, medical imaging~\citep{1250173}, and emotional-behavior analysis~\citep{10494076}. However, traditional image annotation tasks primarily rely on manual processes for selecting objects and assigning labels~\citep{zhang2012review}. This approach is not only time- and labor-intensive~\citep{zhang2012review} but also prone to causing annotator fatigue due to prolonged repetitive work, which in turn affects the quality and consistency of the annotations~\citep{9650877}.

\vspace{-0.1cm}

To alleviate the burden of manual annotation, crowdsourcing methods have been widely adopted. For example, by leveraging Completely Automated Public Turing test to tell Computers and Humans Apart (CAPTCHAs)~\citep{von2003captcha, barnard2003matching, chew2004image} and crowdsourcing platforms, large-scale annotation projects can be decomposed into numerous small tasks that are distributed to participants worldwide, with each participant responsible for annotating only a subset of images~\citep{luz2015survey}. This approach not only effectively reduces costs and improves efficiency but also enhances annotation accuracy through multiple rounds of verification and the collective intelligence of the crowd~\citep{10.1145/1743384.1743478}. Crowdsourcing is generally well-suited for simple, highly repetitive, and standardized tasks, such as basic image annotation or classification (e.g., the CAPTCHAs commonly seen on websites' login page). However, due to limitations in human knowledge, cultural differences, and experience, crowdsourcing is not suitable for generating more detailed labels (as illustrated in Figure 2, which simulates tasks that may be beyond the capabilities of non-experts)~\citep{nassar2019overview}.

\vspace{-0.1cm}

In addition, automated tools and semi-automated annotation methods have gradually emerged in recent years. By leveraging pre-trained object detection models, semantic segmentation algorithms, and image captioning technologies, some systems attempt to automatically generate candidate bounding boxes and initial labels, with subsequent human correction through a human–machine collaborative process~\citep{zhang2012review,cheng2018survey}. Although these methods have improved annotation efficiency to some extent, they still suffer from limitations in generalization and user interaction. Existing automated tools often rely on specific datasets and scenarios, and when faced with diverse or complex situations, models are prone to misclassification and may struggle to accurately capture all details~\citep{10540016}.

\vspace{-0.1cm}

Recently, large language models (LLMs) and large multimodal models (LMMs) such as GPT have achieved groundbreaking progress in the field of natural language processing, and their exceptional semantic understanding and generation capabilities have provided new perspectives for annotation tasks~\citep{doi:10.1073/pnas.2305016120,belal2023leveragingchatgpttextannotation,zhang2024redefiningqualitativeanalysisai,10534765,10.1145/3589334.3648141,10.1145/3589334.3645642,10.1145/3613904.3642834,tan2024largelanguagemodelsdata,10.1145/3627508.3638322,10.1145/3613904.3641960,li2024laionsgenhancedlargescaledataset,zhang2023qualigptgpteasytousetool,lu2023humanwinsllmempirical,zhang2025benchmarkingzeroshotfacialemotion}. This paper proposes a human–AI collaborative annotation framework (as shown in Figure~\ref{fig.framework}), where humans are responsible for selecting target regions in images while an LMM automatically generates labels that align with the image context. The framework has the following two main advantages: (1) reduced human workload by delegating the laborious task of label assignment to the LMM, human annotators can focus solely on target selection, thereby significantly enhancing overall efficiency; and (2) bidirectional human–AI alignment in terms of knowledge and annotation accuracy. From the human annotators’ perspective, they provide guidance (through the selection of regions) to help the LMM more effectively address specific task objectives and use their expertise to verify, correct, and offer feedback on the labels generated by the LMM. From the LMM’s perspective, it can offer more detailed labels to compensate for the human annotators’ potential lack of domain-specific knowledge. Furthermore, by leveraging the LMM's visual analysis capabilities, human annotators are not confined to the limitations of the annotation task, enabling them to broaden the scope of the task. 
%\vspace{-1.2em}
\vspace{-0.2cm}
\section{Proposed Framework}

\vspace{-0.2cm}

%\begin{figure}[h!]
%\centering
%\includegraphics[width=0.9\textwidth]{graph/method-in-this-paper.png}
%\caption{Image Annotation Workflow. The figure illustrates the steps involved in the image annotation process. It begins with a collection of original images, followed by human selection of relevant images. Selected images are then annotated with bounding boxes to highlight objects of interest. The final output consists of labeled bounding boxes, which are used for downstream tasks in computer vision. The rightmost part indicates the task levels that annotators with different knowledge levels can complete. A single asterisk (*) marks labels that can be annotated by the average person. Double asterisks (**) mark labels that can be annotated by those with some foundational or passing-knowledge. Triple asterisks (***) signify labels that only expert groups are deemed capable of annotating (Note: GPT-4o can annotate at this level).}
%\label{fig.framework}
%\end{figure}

\begin{wrapfigure}{r}{0.8\columnwidth}
  \centering
  \includegraphics[width=0.8\columnwidth]{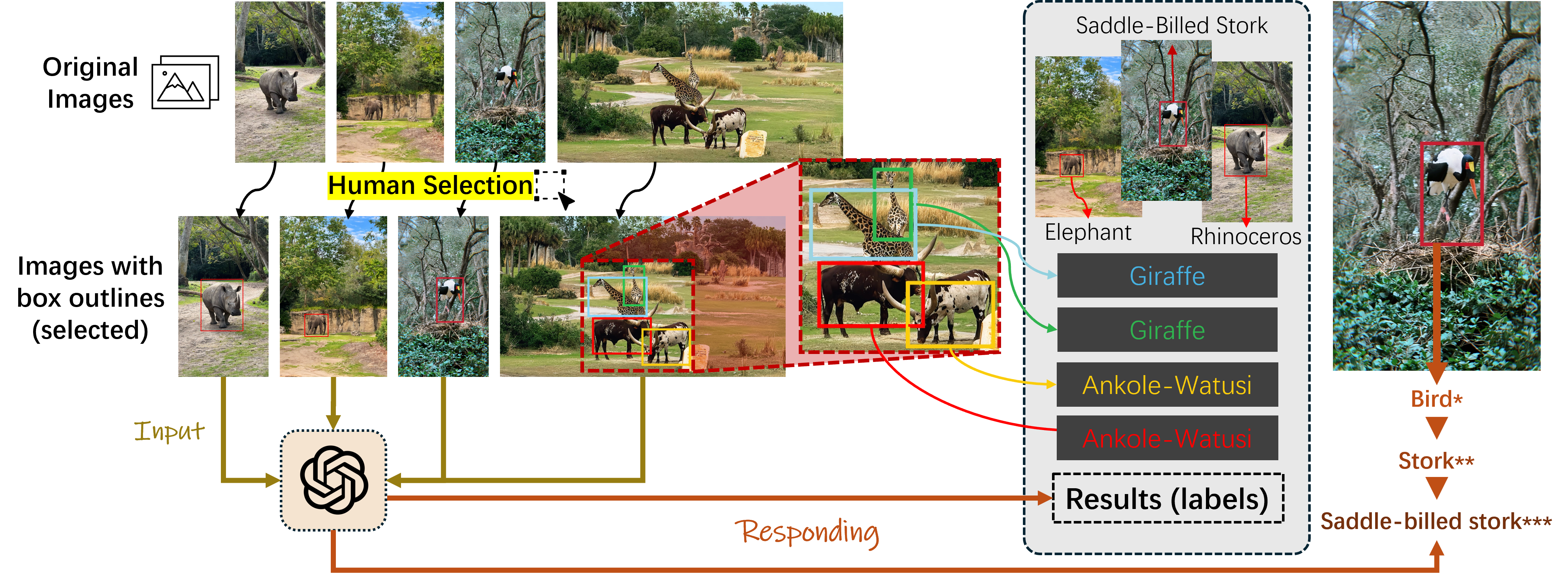}
  \caption{Image Annotation Workflow. The figure illustrates the steps involved in the image annotation process. It begins with a collection of original images, followed by human selection of relevant images. Selected images are then annotated with bounding boxes to highlight objects of interest. The final output consists of labeled bounding boxes, which are used for downstream tasks in computer vision. The rightmost part indicates the task levels that annotators with different knowledge levels can complete. A single asterisk (*) marks labels that can be annotated by the average person. Double asterisks (**) mark labels that can be annotated by those with some foundational or passing-knowledge. Triple asterisks (***) signify labels that only expert groups are deemed capable of annotating (Note: GPT-4o can annotate at this level).}
  \label{fig.framework}
\end{wrapfigure}

Our proposed framework (shown in Figure~\ref{fig.framework}) streamlines the image annotation process through a systematic workflow.  The process begins with a collection of raw images containing various objects of interest. Human annotators then review these images and draw bounding boxes around target objects, helping AI focus on the target and establishes connections between objects and the possible labels. These annotated images are then processed by a LMM, which analyzes the content within each bounding box using prompts such as ``\textit{Please tell me what is selected by the bounding box in each image.}'' The LMM leverages its natural language understanding capabilities to generate precise labels for the outlined objects. This approach yields specific, high-quality annotations, for example, identifying specimens like ``Saddle-Billed Stork'', ``Elephant Rhinoceros'', ``Giraffe'', and ``Ankole-Watusi'', that serve as valuable input for downstream tasks such as object recognition or classification. If an image does not have a bounding box, the LMM will analyze the entire image. In some cases where the image contains only a single subject, a bounding box might not be necessary.
\vspace{-0.2cm}
%\section{Testing and Results}

\subsection{Comparison of the Traditional Workflow and the LMM-Enhanced Annotation Workflow}

\vspace{-0.2cm}

%\begin{figure}[h!]
%\centering
%\includegraphics[width=0.8\textwidth, trim=0 5 0 0, clip]{graph/method-between-human-and-ai.png}
%\caption{Comparison of the Traditional Workflow (left side) and the LMM-Enhanced Annotation Workflow (right side). The workflows are processed from top to bottom.}
%\label{fig.method-between-human-and-ai}
%\end{figure}

\begin{wrapfigure}{r}{0.65\columnwidth}
  \centering
  \includegraphics[width=0.65\columnwidth]{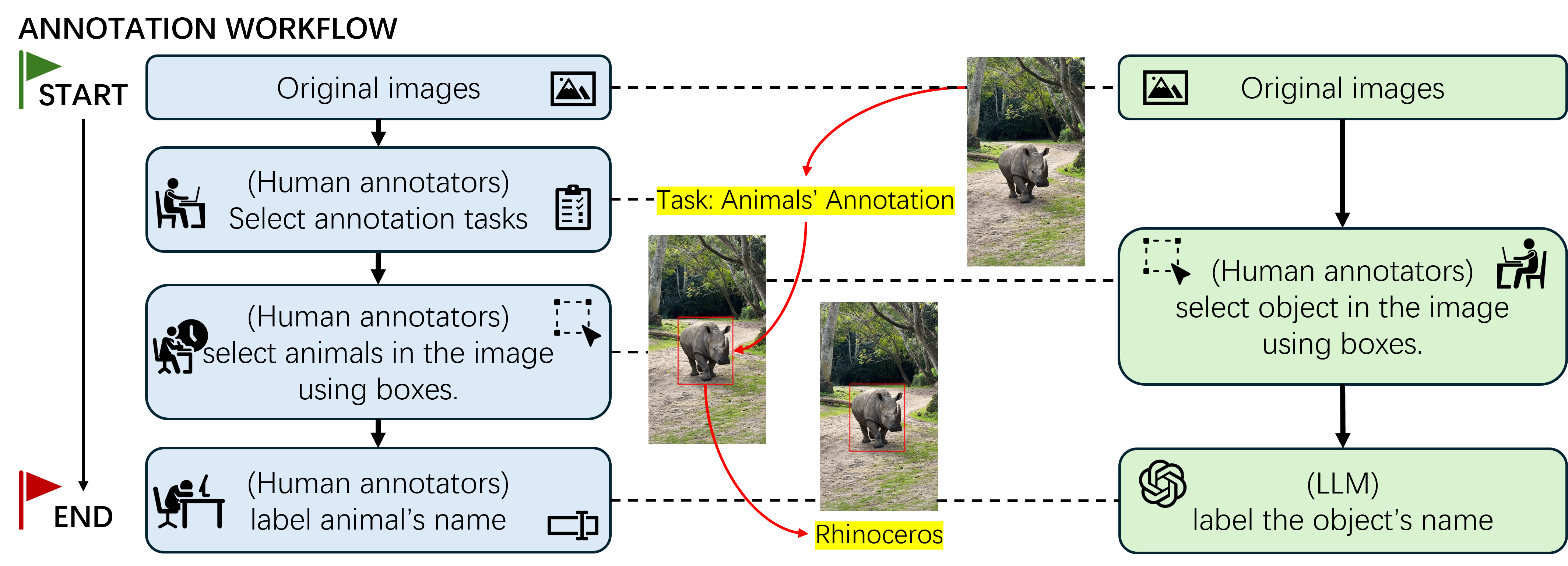}
  \caption{A Synergistic Loop Illustrating Bidirectional Human–AI Alignment.}
  \label{fig.method-between-human-and-ai}
\end{wrapfigure}

Figure~\ref{fig.method-between-human-and-ai} compares the workflow of traditional annotation tasks with the framework proposed in this paper. The traditional workflow places the entire burden of annotation on human annotators, who must both select objects and assign labels. These annotators first choose a specific predefined task, such as identifying animals, and maintain this focus throughout the process. They carefully draw bounding boxes around relevant objects, a step that demands precision since box accuracy directly influences annotation quality. The annotators then assign labels to each bounded region, a task that often requires specialized knowledge. When identifying animal species, for instance, annotators must navigate challenges such as blurry images or complex scenes. While this method can yield high-quality results, it suffers from three key limitations: heavy labor requirements, potential inconsistencies, and poor scalability.
\vspace{-0.1cm}

\vspace{-0.1cm}

The LMM-enhanced workflow addresses these limitations by dividing responsibilities between humans and machines. Human annotators now focus solely on drawing bounding boxes around objects of interest, without restricting themselves to specific categories. This approach reduces cognitive load while establishing the necessary context for subsequent machine processing. The LMM then analyzes these bounded regions and generates appropriate labels based on contextual prompts. In the example case of animal identification, the LMM can supply precise species names without requiring specialized knowledge from human annotators. This hybrid approach offers several advantages over the traditional workflow. By delegating the classification task to LMMs, it reduces the need for specialized expertise, lowers annotation costs, and significantly improves scalability. The division of labor between human visual expertise and machine classification capabilities creates a more efficient and sustainable annotation process, particularly for large datasets.

\vspace{-0.2cm}
\subsection{Testing and Results}
\vspace{-0.2cm}
To validate our LMM-enhanced annotation approach, we extended our evaluation to the Asirra dataset~\citep{10.1145/1315245.1315291}\footnote{Testing utilized the annotated version available on Kaggle, \url{https://www.kaggle.com/datasets/alvarole/asirra-cats-vs-dogs-object-detection-dataset}}, employing GPT-4-mini for rapid annotation testing\footnote{Given the predominantly single-subject nature of the images, we omitted bounding box selection. Additionally, we utilized basic functionality prompts}. The results revealed remarkable accuracy in primary classification tasks, achieving a 99.63\% success rate in distinguishing between cats and dogs.
Beyond basic classification, the proposed framework demonstrated sophisticated labeling capabilities. The LMM successfully generated detailed breed-specific annotations, such as "Dachshund (Dog)", "German Shepherd (Dog)", "Siamese cat (Cat)", and "Himalayan cat (Cat)". This granular classification ability highlights the system's potential for specialized annotation tasks that traditionally require expert knowledge.
These results underscore two key advantages of our approach: exceptional accuracy in basic classification tasks and the ability to provide detailed, breed-specific labels without additional human expertise. This combination of high accuracy and detailed classification capabilities suggests that LMM-enhanced annotation systems can effectively bridge the gap between efficiency and annotation depth.

\vspace{-0.4cm}

\section{Framework Supports Bidirectional Human-AI Alignment}
\vspace{-0.1cm}

\begin{wrapfigure}{r}{0.38\textwidth}
  \centering
  \includegraphics[width=0.38\textwidth, trim=0 2 0 0, clip]{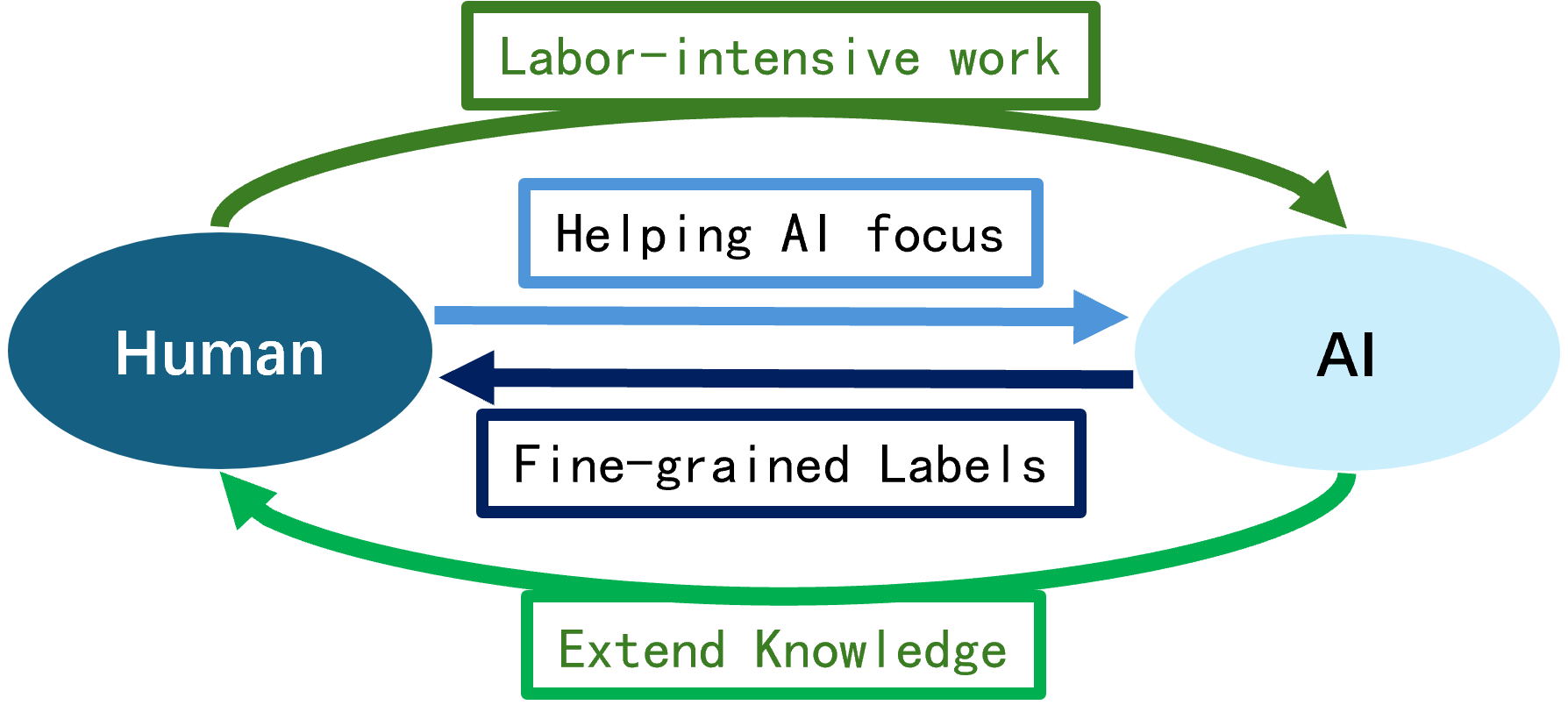}
  \caption{A Synergistic Loop Illustrating Bidirectional Human–AI Alignment.}
  \label{fig.bi-alignment}
\end{wrapfigure}
\vspace{-0.2cm}
This framework emphasizes bidirectional alignment~\citep{shen2024bidirectionalhumanaialignmentsystematic} between human annotators and AI systems, specifically examining the interplay of labor distribution, knowledge transfer, and collaboration. On the one hand, from the human perspective, annotators provide essential guidance by selecting objects of interest through bounding boxes, helping LMMs focus on relevant areas and generate accurate labels. This division of labor reduces the cognitive burden on human annotators while LMMs complement potential gaps in human knowledge by generating detailed, contextually relevant labels, particularly in domains requiring specialized expertise like fine-grained animal species identification. On the other hand, from the AI perspective, LMMs benefit from the structured input provided by human annotators, enhancing their ability to understand and interpret visual content. Through prompt engineering and human selection, LMMs align their outputs with task objectives. This bidirectional interaction not only improves annotation accuracy but also creates a collaborative environment where human and AI systems learn from each other, forming a synergistic loop (as shown in Figure~\ref{fig.bi-alignment}). Over time, this alignment leads to more robust and adaptable annotation systems capable of handling complex tasks across diverse domains.
%\begin{figure}[h!]
%\centering
%\includegraphics[width=0.5\textwidth]{graph/bi-alignment.png}
%\caption{A Synergistic Loop Illustrating Bidirectional Human–AI Alignment.}
%\label{fig.bi-alignment}
%\end{figure}

\vspace{-0.3cm}

\section{Future Work: Data Explosion and Endless Annotation}
\vspace{-0.3cm}
In recent years, the exponential growth of digital data has spawned what's commonly called a ``Data Explosion/Information Explosion.''~\citep{turi2024data,sweeney2001information} As information proliferates across sectors, the demand for annotated datasets has surged to train and maintain high-performance AI models~\citep{liang2022advances,10.1145/3529755}. Traditional annotation methods, heavily reliant on manual labor, struggle to keep pace with this relentless influx, creating a seemingly endless annotation backlog.

\vspace{-0.1cm}

Our framework accelerates the annotation process by leveraging LMM capabilities. By shifting repetitive, labor-intensive tasks to AI, human annotators can focus on critical decisions like object selection and quality validation. Furthermore, it is essential to consider the economic and ethical implications of this approach. From an economic perspective, using LMM for annotation can undoubtedly reduce substantial labor costs, but it also raises further demands for computing resources~\citep{bhattacharya2024demystifying}. This trade-off can be analyzed in future work by comparing the savings from reduced human labor against the costs associated with using the LMM's API, potentially providing a clearer understanding of sustainability and return on investment. Additionally, deploying LMM locally and utilizing smaller-scale models might further reduce costs. From an ethical perspective, although automated annotation can improve efficiency, it also raises concerns about job cuts~\citep{zarifhonarvar2024economics}. Future work could focus on mitigating the negative impact on human workers, possibly by redefining their roles in the annotation workflow and directing them toward more strategic, high-level tasks. This balanced approach not only advances technological progress but also addresses broader societal impacts.

\vspace{-0.1cm}

Despite these improvements, maintaining annotation quality across expanding datasets remains challenging. Future research could explore ways to enhance our framework's scalability. For instance, integrating active learning techniques~\citep{prince2004does} could help the system prioritize the most informative samples, optimizing both human and AI efforts. Additionally, employing image segmentation techniques~\citep{han2024deep} to replace reliance on manual object boxing could enable the framework to operate more autonomously, adapting to new tasks and domains with minimal human intervention, ultimately transferring the endless annotation tasks brought by the data explosion entirely to AI. 
\vspace{-0.3cm}
\subsubsection*{Acknowledgments}
\vspace{-0.2cm}
We would like to extend our gratitude to Wen Chen and Xiaomeng Li for their assistance with the preliminary step of this work. We appreciate the insightful comments and suggestions provided by the anonymous reviewers. This work was supported by the Beijing Natural Science Foundation-Youth Project (Grant No.4254082).

\bibliography{iclr2025_conference}
\bibliographystyle{iclr2025_conference}

%\appendix
%\section{Appendix}
%You may include other additional sections here.

\end{document}